\documentclass{bmvc2k}

\title{Sketch-a-Net that Beats Humans}

\addauthor{Qian Yu$^*$}{q.yu@qmul.ac.uk}{1}
\addauthor{Yongxin Yang$^*$}{yongxin.yang@qmul.ac.uk}{1}
\addauthor{Yi-Zhe Song}{yizhe.song@qmul.ac.uk}{1}
\addauthor{Tao Xiang}{t.xiang@qmul.ac.uk}{1}
\addauthor{Timothy M. Hospedales}{t.hospedales@qmul.ac.uk}{1}

\addinstitution{
School of Electronic Engineering and Computer Science\\
Queen Mary, University of London\\
London, E1 4NS\\
United Kingdom}

\runninghead{Yu, Yang, Song, Xiang, Hospedales}{Sketch-a-Net that Beats Humans}

\makeatletter
\def\blfootnote{\gdef\@thefnmark{}\@footnotetext}
\makeatother

\begin{document}

\maketitle

\begin{abstract}
We propose a multi-scale multi-channel deep neural network framework that, for the first time, yields sketch recognition performance surpassing that of humans. Our superior performance is a result of explicitly embedding the unique characteristics of sketches in our model: (i) a network architecture designed for sketch rather than natural photo statistics, (ii)  a multi-channel generalisation that encodes sequential ordering in the sketching process, and (iii) a multi-scale network ensemble with joint Bayesian fusion that accounts for the different levels of abstraction exhibited in free-hand sketches. We show that state-of-the-art deep networks specifically engineered for photos of natural objects fail to perform well on sketch recognition, regardless whether they are trained  using photo or sketch. Our network on the other hand not only delivers the best performance on the largest human sketch dataset to date, but also is  small in size making efficient training possible using just CPUs.
\end{abstract}

\section{Introduction}
\label{sec:intro}
Sketches are very intuitive to humans and have long been used as an effective communicative tool.  With the proliferation of touchscreens, sketching has become a much easier undertaking for many -- we can sketch on phones, tablets and even watches. Research on sketches has consequently flourished in recent years, with a wide range of applications being investigated, including sketch recognition \cite{eitz2012hdhso, Schneider:2014:SCC:2661229.2661231}, sketch-based image retrieval \cite{eitz2011sbir, rui2013}, sketch-based 3D model retrieval \cite{sketchecvpr15}, and forensic sketch analysis \cite{Klare11matchingforensic,OuYangHSL14}.\blfootnote{\hspace{-1.9em}$^*$ These authors contributed equally to this work}

Recognising free-hand sketches (e.g.~asking a person to draw a car without any instance of car as reference) is an extremely challenging task. This is due to a number of reasons: (i) sketches are highly iconic and abstract, e.g., human figures can be depicted as stickmen; (ii) due to the free-hand nature, the same object can be drawn with hugely varied levels of detail/abstraction, e.g., a human figure sketch can be either a stickman or a portrait with fine details depending on the drawer; (iii) sketches lack visual cues, i.e., they consist of black and white lines instead of coloured pixels.  A recent large-scale study on 20,000 free-hand sketches across 250 categories of daily objects puts human sketch recognition accuracy at 73.1\% \cite{eitz2012hdhso}, suggesting that the task is challenging even for humans.

Prior work on sketch recognition generally follows the conventional image classification paradigm, that is, extracting hand-crafted features from sketch images followed by feeding them to a classifier. Most hand-crafted features traditionally used for photos (such as HOG, SIFT and shape context) have been employed, which are often coupled with Bag-of-Words (BoW) to yield a final feature representations that can then be classified. 
However, existing hand-crafted features designed for photos do not account for the unique abstract and sparse nature of sketches. Furthermore,  they ignore a  key unique characteristics of sketches, that is, a sketch is essentially an ordered list of strokes; they are thus sequential in nature. In contrast with photos that consist of pixels sampled all at once, a sketch is the result of an online drawing process. It had long been recognised in psychology \cite{johnson2009computational} that such sequential ordering is a strong cue in human sketch recognition, a phenomenon that is also confirmed by recent studies in the computer vision literature \cite{Schneider:2014:SCC:2661229.2661231}. However, none of the existing approaches attempted to embed sequential ordering of strokes in the recognition pipeline even though that information is readily available. 

In this paper, we propose a novel deep neural network (DNN), Sketch-a-Net, for free-hand sketch recognition, which is specifically designed to accommodate the unique characteristics of sketches including multiple levels of abstraction and being sequential in nature. DNNs, especially deep convolutional neural networks (CNNs) have achieved tremendous successes recently in replacing representation hand-crafting with representation learning  for a variety of vision problems \cite{NIPS2012_4824, Simonyan14c}. However, existing DNNs are primarily designed for photos; we demonstrate experimentally that directly employing them for the sketch modelling problem produces little improvement over hand-crafted features, indicating special model architecture is required for sketches. To this end, our Sketch-a-Net has three key features that distinguish it from the existing DNNs: (i) a number of model architecture and learning parameter choices specifically for addressing the iconic and abstract nature of sketches;  (ii) a multi-channel architecture designed to model the sequential ordering of strokes in each sketch; and (iii) a multi-scale network ensemble to address the variability in abstraction and sparsity, followed by a joint Bayesian fusion scheme to exploit the complementarity of different scales.  The overall model is small in size, being 7 times smaller than the classic AlexNet \cite{NIPS2012_4824} in terms of the number of parameters, therefore making it efficient to train independently of special hardware, i.e.~GPUs. 

Our contributions are summarised as follows: (i) for the first time, a representation learning model based on DNN is presented for sketch recognition in place of the conventional hand-crafted feature based sketch representations; (ii) we demonstrate how  sequential ordering information in sketches can be embedded into the DNN architecture  and in turn improve sketch recognition performance; (iii) we propose a multi-scale network ensemble that fuses networks learned at different scales together via joint Bayesian fusion to address the variability of levels of abstraction in sketches. Extensive experiments on the largest hand-free sketch benchmark dataset, the TU-Berlin sketch dataset \cite{eitz2012hdhso}, show that our model significantly outperforms existing approaches and  can even  beat humans at sketch recognition.  

\section{Related Work}
\label{sec:lit}
\textbf{Free-hand Sketch Recognition:} Early studies on sketch recognition worked with professional CAD or artistic drawings as input \cite{Lu:2005:NRM:1649592.1649930,Jabal2009ICIME, Zitnick:2013:BSF:2514950.2516226,Sousa:2009:GMC:1507775.1508004}. Free-hand sketch recognition had not attracted much attention until very recently when a large crowd-sourced dataset was published in \cite{eitz2012hdhso}. Free-hand sketches are drawn by non-artists using touch sensitive devices rather than purpose-made equipments; they are thus often highly abstract and  exhibit large intra-class deformations. Most existing works \cite{eitz2012hdhso,Schneider:2014:SCC:2661229.2661231,Li2015} use SVM as the classifier and differ only in what hand-crafted features borrowed from photos are used as representation.  Li et al. \cite{Li2015} demonstrated that fusing different local features using multiple kernel learning helps improve the recognition performance. They also examined the performance of many features individually and found that HOG generally outperformed others. Very recently, Schneider and Tuytelaars \cite{Schneider:2014:SCC:2661229.2661231} demonstrated that Fisher Vectors, an advanced feature representation scheme successfully applied to image recognition, can be adapted to sketch recognition and achieve near-human accuracy (68.9\% vs.~73.1\% for humans on the TU-Berlin sketch dataset). 

Despite these great efforts, no attempt was made thus far for either designing or learning  feature representations specifically for sketches. Moreover, the role of sequential  ordering in sketch recognition remains unaddressed. In this paper, we turn to DNNs which have shown great promise in many areas of computer vision \cite{NIPS2012_4824, Simonyan14c} for representation learning. Our learned representation uniquely exploits the sequential ordering information of strokes in a sketch and is able to cope with multiple levels of abstraction in the same sketch category. Note that the optical character recognition (OCR) community has exploited stroke ordering with some success \cite{Yin2013}, yet the problem of encoding sequential information is harder on sketches -- handwriting characters have relatively fixed structural ordering therefore simple heuristics often suffice; sketches on the the other hand exhibit a much higher degree of intra-class variation in stroke ordering, which motivates us to resort to the powerful DNNs to learn the most suitable sketch representation.


\noindent\textbf{DNNs for Visual Recognition:} Deep Neural Networks (DNNs) have recently achieved impressive performance for many recognition tasks across different disciplines. In particular, Convolutional Neural Networks (CNNs) have dominated top benchmark results on visual recognition challenges such as ILSVRC \cite{imagenet_cvpr09}. 
When first introduced in the 1980s, CNNs were the preferable solution for small problems only (e.g. LeNet \cite{Cun90handwrittendigit} for handwritten digit recognition). Their practical applications were severely bottlenecked by the high computational cost when the number of classes and training data are large. However with the recent proliferation of modern GPUs, this bottleneck has been largely alleviated. Nonetheless, it was not until the introduction of ReLU neurons (instead of TanH), max-pooling (instead of average pooling) and dropout regularisation that DNNs maximised their effectiveness and regained their popularity \cite{NIPS2012_4824}. An important advantage of DNNs, particularly CNNs, compared with conventional classifiers such as SVMs, lies with the closely coupled nature of  presentation learning and classification (i.e., from raw pixels to class labels in a single network), which makes the learned feature representation maximally discriminative. More recently, it was shown that even deeper networks with smaller filters \cite{Simonyan14c} are preferable for photo image recognition. Despite these great strides, to the best of our knowledge, all existing image recognition DNNs are optimised for photos, ultimately making them perform sub-optimally on sketches. In this paper, we show that directly applying  successful photo-oriented DNNs to sketches leads to little improvement over hand-crafted feature based methods. In contrast, by embedding the unique characteristics of sketches into the network design, our Sketch-a-Net  advances sketch recognition to the over-human level.

\section{Methodology}
\label{sec:meth}
\begin{figure}[t]
\centering
\includegraphics[width=0.98\textwidth]{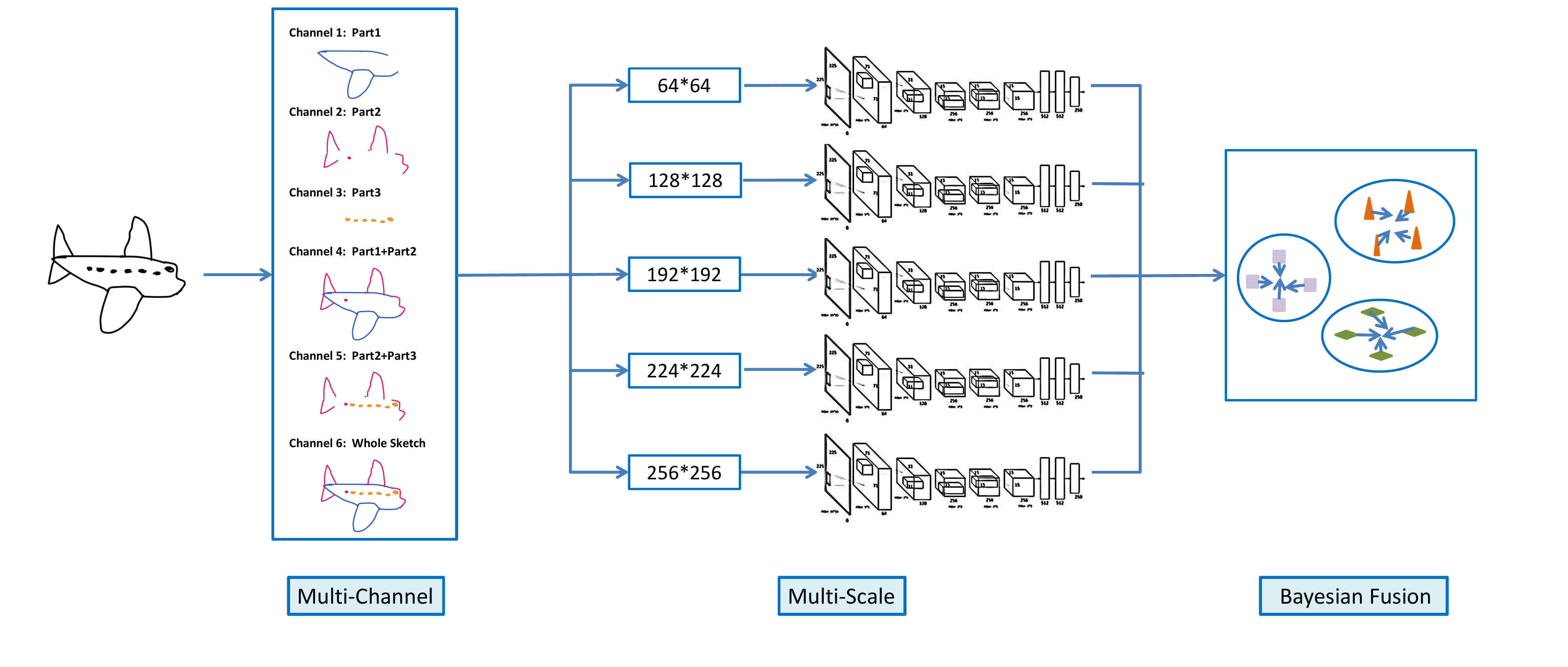}
\caption{Illustration of our  overall framework.}
\label{fig:framework}
\end{figure}

In this section we introduce the three key technical components of our framework. We first detail our basic CNN architecture and outline the important considerations for Sketch-a-Net compared to the conventional photo-oriented DNNs (Sec.~\ref{sec:DNN_basic}). We next explain our simple but novel generalisation that gives a DNN the ability to exploit the stroke ordering information that is unique to sketches (Sec.~\ref{sec:DNN_6c}). We then introduce a multi-scale ensemble of networks to address the variability in the levels of abstraction with a joint Bayesian fusion method for exploiting the complementarity of different scales (Sec.~\ref{sec:Ensemble}). Fig.~\ref{fig:framework} illustrates our overall framework.

\vspace{-0.1cm}
\subsection{A CNN for Sketch Recognition}\label{sec:DNN_basic}
Our Sketch-a-Net is a deep CNN. Despite all the efforts so far, it remains an open question how to design the architecture of CNNs given a specific visual recognition task; but most recent recognition networks \cite{Chatfield14,Simonyan14c} now follow a design pattern  of multiple convolutional layers followed by fully connected layers, as popularised by the work of \cite{NIPS2012_4824}. 

Our specific architecture is as follows: first we use five convolutional layers, each with rectifier (ReLU) \cite{lecun-98b} units, while the first, second and fifth layers are followed by max pooling (Maxpool). The filter size of the sixth convolutional layer (index 14 in Table~\ref{tab:architecture}) is $7\times7$, which is the same as the output from previous pooling layer, thus it is precisely a fully-connected layer. Then two more fully connected layers are appended. Dropout regularisation \cite{hinton2012improving} is applied on the first two fully connected layers. The final layer has 250 output units corresponding to 250 categories (that is the number of unique classes in the TU-Berlin sketch dataset), upon which we place a softmax loss. The details of our CNN are summarised in Table~\ref{tab:architecture}. Note that for simplicity of presentation, we do not explicitly distinguish fully connected layers from their convolutional equivalents.


Most  CNNs are proposed without explaining why  parameters, such as filter size, stride, filter number, padding and pooling size, are chosen. Although it is  impossible to exhaustively verify the effect of every free (hyper-)parameter,  we discuss some points that are consistent with  classic designs, as well as those that are specifically designed for sketches, thus considerably different from the CNNs targeting photos, such as AlexNet \cite{NIPS2012_4824} and DeCAF \cite{Donahue_ICML2014}.

\vspace{0.2cm}\noindent\textbf{Commonalities between Sketch-a-Net and Photo-Oriented CNN Architectures}

\noindent\textbf{Filter Number:}\quad In both our Sketch-a-Net and recent photo-oriented CNNs \cite{NIPS2012_4824,Simonyan14c}, the number of filters increases with depth. In our case the first layer is set to $64$, and this is doubled after every pooling layer (indicies: $3\rightarrow 4$, $6\rightarrow 7$ and $13\rightarrow 14$) until $512$.

\begin{table}[t]
\centering
\scriptsize
\begin{tabular}{c c c c c c c c c l }
\hline Index & Layer & Type & Filter Size & Filter Num & Stride & Pad & Output Size\\ 
\hline 
0 & & Input & - & - & - & - & $225\times 225$ \\\hline 
1 & L1 & Conv & $15\times 15$ & 64 & 3 & 0 & $71\times 71$ \\
2 & & ReLU & - & - & - & - & $71\times 71$\\
3 & & Maxpool & $3\times 3$ & - & 2 & 0 & $35\times 35$ \\\hline 
4 & L2 & Conv & $5\times 5$ & 128 & 1 & 0 & $31\times 31$ \\
5 & & ReLU & - & - & - & - & $31\times 31$\\
6 & & Maxpool & $3\times 3$ & - & 2 & 0 & $15\times 15$ \\\hline 
7 & L3 & Conv & $3\times 3$ & 256 & 1 & 1 & $15\times 15$\\
8 & & ReLU & - & - & - & - & $15\times 15$\\ 
9 & L4 & Conv & $3\times 3$ & 256 & 1 & 1 & $15\times 15$\\
10 & & ReLU & - & - & - & - & $15\times 15$\\ 
11 & L5 & Conv & $3\times 3$ & 256 & 1 & 1 & $15\times 15$\\
12 & & ReLU & - & - & - & - & $15\times 15$\\
13 & & Maxpool & $3\times 3$ & - & 2 & 0 & $7\times 7$\\\hline 
14 & L6 & Conv(=FC) & $7\times 7$ & 512 & 1 & 0 & $1\times 1$\\
15 & & ReLU & - & - & - & - & $1\times 1$\\
16 & & Dropout (0.50) & - & - & - & - & $1\times 1$\\\hline 
17 & L7 & Conv(=FC) & $1\times 1$ & 512 & 1 & 0 & $1\times 1$\\
18 & & ReLU & - & - & - & - & $1\times 1$\\
19 & & Dropout (0.50) & - & - & - & - & $1\times 1$\\\hline 
20 & L8 & Conv(=FC) & $1\times 1$ & 250 & 1 & 0 & $1\times 1$\\
\hline
\end{tabular}
\vspace{0.2cm}
\caption{The architecture of Sketch-a-Net.} \label{tab:architecture}
\end{table}

\noindent\textbf{Stride:}\quad As with photo-oriented CNNs, the stride of convolutional layers after the first is set to one. This keeps as much information as possible.

\noindent\textbf{Padding:}\quad  Zero-padding is used only in L3-5 (Indices $7$, $9$ and $11$). This is to ensure that the output size is an integer number, as in photo-oriented CNNs \cite{Chatfield14}.

\vspace{0.2cm}
\noindent\textbf{Unique Aspects in our Sketch-a-Net Architecture}

\noindent\textbf{Larger First Layer Filters:}\quad The size of filters in the first convolutional layer might be the most sensitive parameter, as all subsequent processing depends on the first layer output. While classic networks use large $11\times11$ filters \cite{NIPS2012_4824}, the current trend of research \cite{ZeilerF14} is moving toward ever smaller filters: very recent \cite{Simonyan14c} state of the art networks have attributed their success in large part to use of tiny $3\times3$ filters. In contrast, we find that larger filters are more appropriate for sketch modelling. This is because sketches lack texture information, e.g., a small round-shaped patch can be recognised as eye or button in a photo based on texture, but this is  infeasible for sketches. Larger filters thus help to capture more structured context rather than textured information. To this end,  we use a filter size of $15\times15$.   

\vspace{0.2cm}\noindent\textbf{No Local Response Normalisation:}\quad  Local Response Normalisation (LRN) \cite{NIPS2012_4824} implements a form of lateral inhibition, which is found in real neurons. This is used pervasively in contemporary CNN recognition architectures  \cite{NIPS2012_4824,Chatfield14,Simonyan14c}. However, in practice LRN's benefit is due to providing ``brightness normalisation''. This is not necessary in sketches since brightness is not an issue in line-drawings. Thus removing LRN layers makes learning faster without sacrificing performance.

\vspace{0.2cm}\noindent\textbf{Larger Pooling Size:}\quad Many recent CNNs use $2\times2$ max pooling with stride 2 \cite{Simonyan14c}. It efficiently reduces the size of the layer by 75\% while bringing some spatial invariance. However,  we use the modification: $3\times3$ pooling size with stride 2, thus generating overlapping pooling areas \cite{NIPS2012_4824}. We found this brings  $\sim1\%$ improvement without much additional computation. 

\vspace{0.2cm}\noindent\textbf{Higher Dropout:}\quad Deeper neural networks generally improve performance but risk overfitting \cite{Simonyan14c}. Recent CNN successes \cite{Simonyan14c,NIPS2012_4824,Chatfield14} deal with this using the (very large) ImageNet dataset \cite{imagenet_cvpr09} for training, and dropout \cite{hinton2012improving} regularisation (randomly setting units activation to zero). Since a sketch dataset is typically much smaller than ImageNet, we compensate for this by setting a much higher dropout rate of 50\%.




\vspace{0.2cm}\noindent\textbf{Lower Computational Cost:}\quad The total number of parameters in  Sketch-a-Net is 8.5 million, 
which is relatively small for modern CNNs. For example, the classic AlexNet \cite{NIPS2012_4824} has 60 million parameters ($7$ times larger), and recent state-of-the-art \cite{Simonyan14c} reaches 144 million.

\subsection{Modelling Sketch Stroke Order with Multiple Channels}
\label{sec:DNN_6c}

\begin{figure}[t]
\centering
\includegraphics[width=0.47\textwidth]{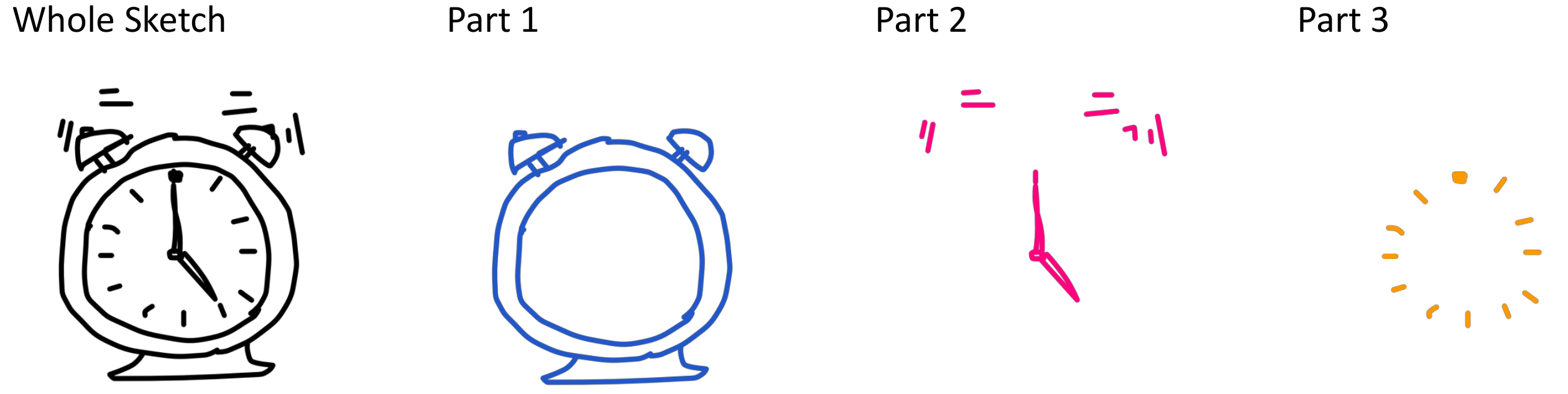} \hspace{0.3cm}
\includegraphics[width=0.47\textwidth]{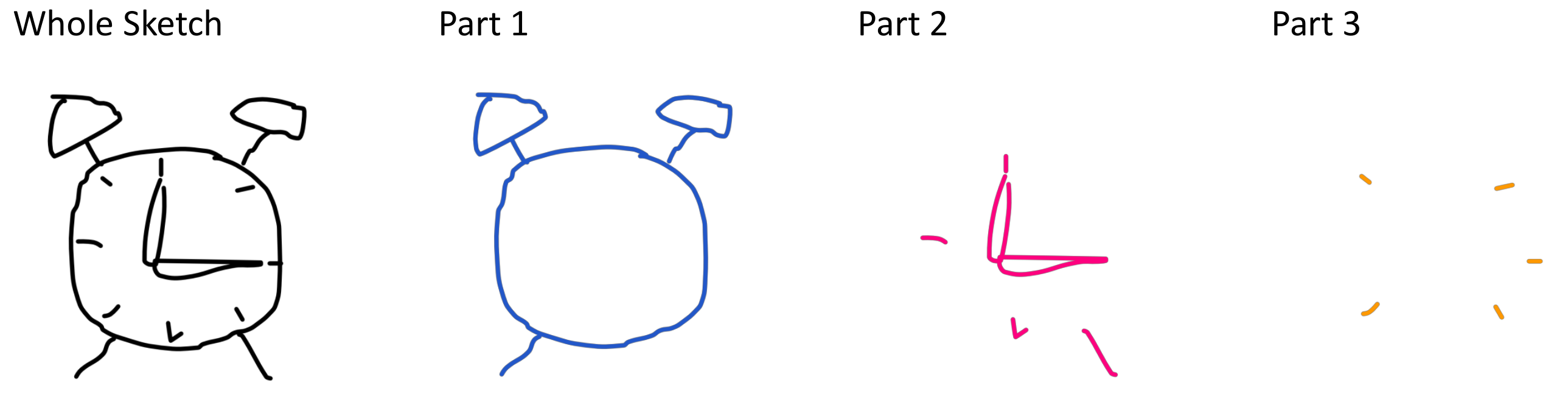}
\caption{Illustration of stroke ordering in sketching with the Alarm Clock category.}
\label{fig:temporal}
\end{figure}

\textbf{Stroke Ordering:}\quad The order of drawn strokes is key information associated with sketches drawn on touchscreens compared to conventional photos where all pixels are captured in parallel. Although this information exists in main sketch datasets such as TU-Berlin, existing work has generally ignored it. To provide intuition about this, Fig.~\ref{fig:temporal} illustrates some sketches in the Alarm Clock category, with strokes broken down into three parts according to stroke order. Clearly there are different sketching strategies in terms of which semantic parts to draw first, but it is common to draw the main outline first, followed by details, as a recent study also found \cite{eitz2012hdhso}.  Modelling stroke ordering information is thus useful in distinguishing categories that are similar in their parts but differ in their typical ordering. 

\vspace{0.2cm}
\noindent \textbf{Modelling Stroke Order:}\quad We propose a simple but effective approach to modelling the sequential order of strokes  by extending Sketch-a-Net to a multi-channel CNN: discretising strokes into three sequential  groups (Fig.~\ref{fig:temporal}), and treating  these parts as different channels in the first layer. Specifically, we use the three stroke parts to generate six images containing  combinations of the stroke parts. As illustrated in Fig.~\ref{fig:framework}, the first three images contain the three parts alone; the next two contain pairwise combinations of two parts, and the third is the original sketch of all parts. Our Sketch-a-Net described in Sec.~\ref{sec:DNN_basic} is then modified to take the six  channel images as input (i.e.~the first layer convolution filter size is changed to $15\times15\times6$). 
This multi-channel model has a couple of advantages: (i) the relative importance of early versus late strokes are learned automatically by back propagation training; (ii) it is a simple and efficient modification of the existing architecture: the number of parameters and hence training time is only increased by $1\%$ compared to the single channel Sketch-a-Net.

\subsection{A Multi-scale Network Ensemble with Bayesian Fusion}\label{sec:Ensemble}
The next challenging aspect of sketch recognition to be addressed is the variability in sketching abstraction. To deal with this we introduce an ensemble of our multi-channel Sketch-a-Nets. For each network in the ensemble we learn a model of varying coarseness by blurring its training data to different degrees. Specifically, we create a 5 network ensemble by blurring -- downsampling and then upsampling by to the original $256\times256$ pixel image size. The downsample sizes are: $256, 224, 192, 128, 64$. Each network in the ensemble is independently trained by backdrop using one of these blur levels. 

The multi-scale Sketch-a-Net  ensemble can be used for classifying a test image using score-level fusion, i.e., averaging the softmax scores.  However, this fusion strategy  treats each network thus each scale equally without discrimination. Alternatively, one could concatenate the CNN learned representations in each network and feed them to a downstream classifier  \cite{Donahue_ICML2014}. However, again no scale and feature selection is possible with this feature-level fusion strategy.  In this work, we propose to take the ($512D$) activation of the penultimate layer of our network as a representation, and apply the recent Joint Bayesian (JB)  fusion method \cite{chen2012jointBayes} to exploit the complementarity between different scales.

The JB framework models  \emph{pairs} of instances (in this case CNN activations), by full covariance Gaussians, under the prior assumption that each instance $x$ is a sum of its (Normally distributed) category mean and instance specific deviation: $x=\mu+\epsilon$. In particular it learns two full covariance Gaussians, representing pairs from the same category and different categories respectively, i.e., it models $p(x_1,x_2|H_I)$ and $p(x_1,x_2|H_E)$ where $x_1$ and $x_2$ are instances, and $H_I$ and $H_E$ are the matched and mismatched pair hypotheses respectively. JB provides an EM algorithm for learning these covariances $\Sigma_I$ and $\Sigma_E$ respectively. Once learned, optimal Bayesian matching can be done using a likelihood ratio test: 

\begin{equation}
r(x_1,x_2)=\text{log}\frac{P(x_1,x_2\mid H_I)}{P(x_1,x_2\mid H_E)}
\label{rx1x2}
\end{equation}

\noindent which turns out to be equivalent \cite{chen2012jointBayes} to a metric learner capable of learning strong metrics with more degrees of freedom than traditional Mahalanobis metrics. 

Although initially designed for \emph{verification}, we re-purpose JB for classification here. Let each $x$ represent the $5\times512=2560D$ concatenated feature vector from our network ensemble. Training: Using this activation vector as a new representation for the training data, we train the JB model, thus learning a good metric. Testing: Given the activation vectors of train and test data, we use the  likelihood-ratio test (Eq.~\ref{rx1x2}) to compare each test point to the full train set. With this mechanism to match test to train points, final classification is achieved with K-Nearest-Neighbour (KNN) matching\footnote{We set $k=5$ in this work and the regularisation parameter of JB is set to $0.1$}\footnote{For robustness at test time, we also take 10 crops and reflections of each train and test image \cite{NIPS2012_4824}. This inflates the KNN train and test pool by 10, and the crop-level matches are combined to image predictions by majority voting.}. Note that in this way each feature dimension from each network is fused together, implicitly giving more weight to more important features, as well as finding the optimal combination of different features at different scales.

\section{Experiments}

\noindent\textbf{Dataset:}\quad We evaluate our model on the TU-Berlin sketch dataset \cite{eitz2012hdhso}, which is the largest and now the most commonly used human sketch dataset. It contains 250 categories with 80 sketches per category. It was collected on Amazon Mechanical Turk (AMT) from 1,350 participants, thus providing a diversity of both categories and sketching styles within each category. We rescaled all images to $256\times256$ pixels in order to make it comparable with previous work. Also following previous work we performed 3-fold cross-validation within this dataset (2 folds for training and 1 for testing).

\vspace{0.2cm}
\noindent\textbf{Data Augmentation:}\quad Data augmentation is commonly with CNNs to reduce overfitting. We  performed data augmentation by replicating the sketches with a number of transformations. Specifically, for each input sketch, we did horizontal reflection, rotation (in the range [-5, +5] degrees) and systematic combinations of horizontal and vertical shifts (up to 32 pixels). Thus, when using two thirds of the data for training, the total pool of training instances is $(20000\cdot0.67)\times(32\cdot32\cdot11\cdot2)=300M$, increasing the  size by a factor of 22,528.

\vspace{0.2cm}
\noindent\textbf{Competitors:}\quad We compared our results with a variety of alternatives. These included the  conventional \textbf{HOG-SVM} pipeline \cite{eitz2012hdhso}, structured \textbf{ensemble matching} \cite{yi2013bmvc}, \textbf{multi-kernel SVM} \cite{Li2015}, the current state-of-the-art Fisher Vector Spatial Pooling \textbf{(FV-SP)} \cite{Schneider:2014:SCC:2661229.2661231}, and  DNN based models including  \textbf{AlexNet} \cite{NIPS2012_4824} and \textbf{LeNet} \cite{lecun-98b}.  AlexNet is a large deep CNN designed for classifying ImageNet LSVRC-2010 \cite{imagenet_cvpr09} images. It has five convolutional layers and 3 fully connected layers. We used two versions of  AlexNet: (i) \textbf{AlexNet-SVM}:  following common practice \cite{Donahue_ICML2014},  it was used as a  pre-trained feature extractor, by taking the second 4096D fully-connected layer of the ImageNet-trained model as a feature vector for SVM classification. (ii) \textbf{AlexNet-Sketch}: we re-trained AlexNet for the 250-category sketch classification task, i.e.~it was trained using  the same data as our  Sketch-a-Net. Finally, although LeNet is quite old, we note that it is specifically designed for handwritten digits rather than photos. Thus it is potentially more suited for sketches than the photo-oriented AlexNet. 

\noindent
\begin{table}[ht!]
\resizebox{\textwidth}{!}{  
\begin{tabular}{ccccccc}
\hline 
HOG-SVM \cite{eitz2012hdhso} & Ensemble \cite{yi2013bmvc} & MKL-SVM \cite{Li2015}& FV-SP \cite{Schneider:2014:SCC:2661229.2661231}  &\tabularnewline
56\% & 61.5\% & 65.8\% & 68.9 &\tabularnewline
AlexNet-SVM \cite{NIPS2012_4824} & AlexNet-Sketch \cite{NIPS2012_4824} & LeNet \cite{lecun-98b} & Sketch-a-Net    & Human \cite{eitz2012hdhso} &\tabularnewline
67.1\% & 68.6\% & 55.2\% & \textbf{74.9\%} &  73.1\% \tabularnewline
\hline 
\end{tabular}
}
\caption{Comparison with state of the art results on sketch recognition}\label{tab:stateoftheart}
\end{table}


\noindent\textbf{Comparative Results:}\quad  We first report the sketch recognition results of our full Sketch-a-Net, compared to state-of-the-art alternatives as well as humans in Table \ref{tab:stateoftheart}. The following observations can be made: (i) Sketch-a-Net significantly outperforms all existing methods  purpose designed for sketch \cite{eitz2012hdhso,yi2013bmvc,Schneider:2014:SCC:2661229.2661231}, as well as the state-of-the-art photo-oriented CNN model \cite{NIPS2012_4824} repurposed for sketch; (ii) we show  for the first time, an automated sketch recognition model can  surpass human performance on  sketch recognition (74.9\% by our Sketch-a-Net vs.~73.1\% for humans based on the study in \cite{eitz2012hdhso}); (iii) Sketch-a-Net is superior to AlexNet, despite being much smaller with only 14\% of the total number of parameters of AlexNet. This verifies that new network design is required for sketch images. In particular, it is noted that either trained using the larger ImageNet data (67.1\%) or the sketch data (68.6\%), AlexNet cannot beat the best hand-crafted feature based approach (68.9\% of FV-SP);  (iv) among the deep DNN based models, the performance of LeNet (55.2\%) is the weakest. Although designed for handwriting digit recognition, a task similar to that of sketch recognition, the model is much simpler and shallow. This suggests that a deeper/more complex model is necessary to cope with the larger intra-class variations exhibited in sketches; (v) last but not least, upon close category-level examination, we found that Sketch-a-Net tends to perform better at fine-grained object categories. This indicates that Sketch-a-Net learned a more discriminative feature representation capturing finer details than conventional hand-crafted features, as well as human. For example, for `seagull', `flying-bird', `standing-bird' and `pigeon', all of which belong to the coarse semantic category of `bird',
Sketch-a-Net obtained an average accuracy of 42.5\% while human only achieved 24.8\%. In particular, the category `seagull', is the worst performing category for human with an accuracy of just 2.5\%, since it was mostly confused with other types of birds. In contrast, Sketch-a-Net yielded 23.9\% for `seagull' which is nearly 10 times better. 

\begin{table}[ht]
\centering
\begin{tabular}{ccccc}
\hline 
Full Model (M-Cha+M-Sca) & M-Cha+S-Sca  & S-Cha+S-Sca & AlexNet-Sketch \tabularnewline
\hline 
74.9\% & 72.6\% & 72.2\% & 68.6\% \tabularnewline
\hline 
\end{tabular}
\protect\caption{Evaluation on the contributions of individual components of Sketch-a-Net. }\label{tab:breakdown}
\end{table}

\noindent\textbf{Contributions of Individual Components:}\quad Compared to conventional photo-oriented DNNs such as AlexNet, our Sketch-a-Net has three distinct features: (i) the specific network architecture (see Sec.~\ref{sec:DNN_basic}), (ii) the multi-channel structure for modelling stroke ordering (see Sec.~\ref{sec:DNN_6c}), and (iii) the multi-scale network ensemble to deal with variable levels of abstraction (see Sec.~\ref{sec:Ensemble}).  In this experiment,  we evaluate the contributions of each new feature.  Specifically, we examined two stripped-down versions of our full model (multi-channel-multi-scale (M-Cha+M-Sca)): multi-channel-single-scale (M-Cha+S-Sca) Sketch-a-Net which uses only one network scale (the original scale of $256\times256$), and single-channel-single-scale (S-Cha+S-Sca) Sketch-a-Net which uses only sketches at the original scale. Results in Table \ref{tab:stateoftheart} show that all three features contribute to the final strong performance of Sketch-a-Net. In particular, (i) the improvement of S-Cha+S-Sca over AlexNet-Sketch shows that our sketch-specific network architecture is effective; (ii) M-Cha+S-Sca achieved better performance than  S-Cha+S-Sca, indicating  the multi-channel features worked; (iii) the best result is achieved when all three new features are combined.

\begin{table}[ht]
\centering 
\begin{tabular}{cccc}
\hline 
Joint Bayesian & Feature Fusion & Score Fusion \tabularnewline
\hline 
74.9\% & 72.8\% & 74.1\% \tabularnewline
\hline 
\end{tabular}
\protect\caption{Comparison of different fusion strategies.}\label{tab:ensemble}
\end{table}

\noindent\textbf{Comparison of Different Fusion Strategies:}\quad Given an ensemble of Sketch-a-Net at different scales, various fusion strategies can be adopted for the final classification task. Table~\ref{tab:ensemble} compares our joint Bayesian fusion method with the two most commonly adopted alternatives: feature level fusion and score level fusion. For feature level fusion, we treat each single scale network as a feature extractor, and concatenate the 512D outputs of their penultimate layers into a single feature. We then trained a linear SVM based on this $5\times512=2560$D feature vector. For score level fusion, we average the $250$D softmax probabilities of each network in the ensemble to make a final prediction. For JB fusion, we take the same $2560$D concatenated feature vector used by feature fusion, but perform KNN matching with JB similarity metric, rather than SVM classification. Interestingly, although score fusion is better than vanilla SVM feature fusion, JB makes much better use of the concatenated feature vector because the full covariance model better learns how to weight the outputs of the networks and exploit their complementarity.

 
\begin{figure}[t]
\centering
\includegraphics[width=1.0\textwidth]{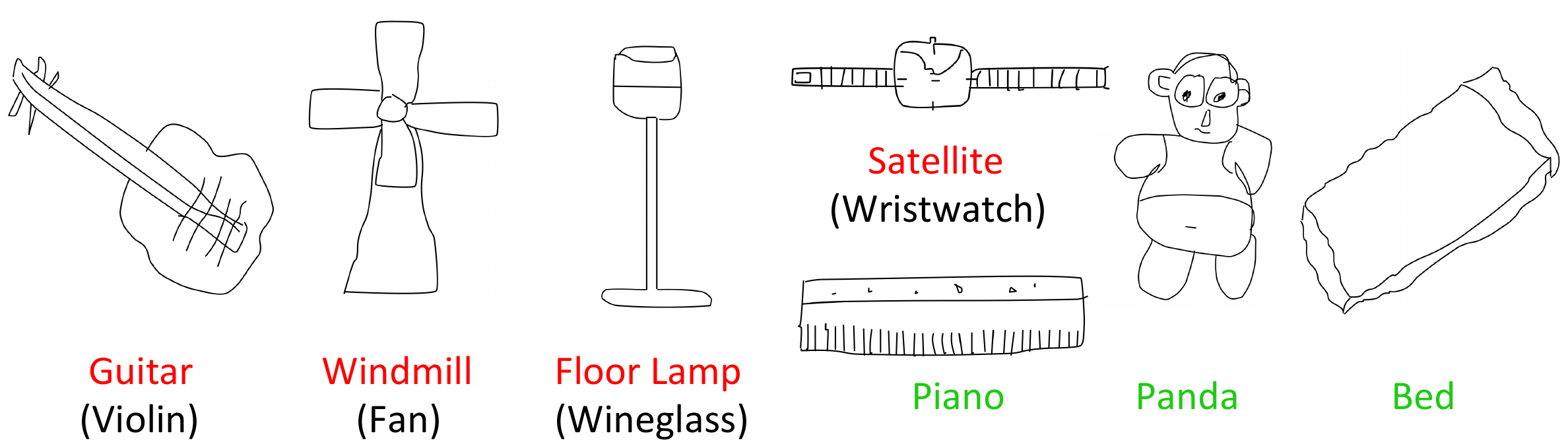} 
\vspace{-0.3cm}
\caption{Qualitative illustration of recognition successes (green) and failures (red).}
\label{fig:qualitative}
\end{figure}

\begin{figure}[ht]
\centering
\includegraphics[width=18em]{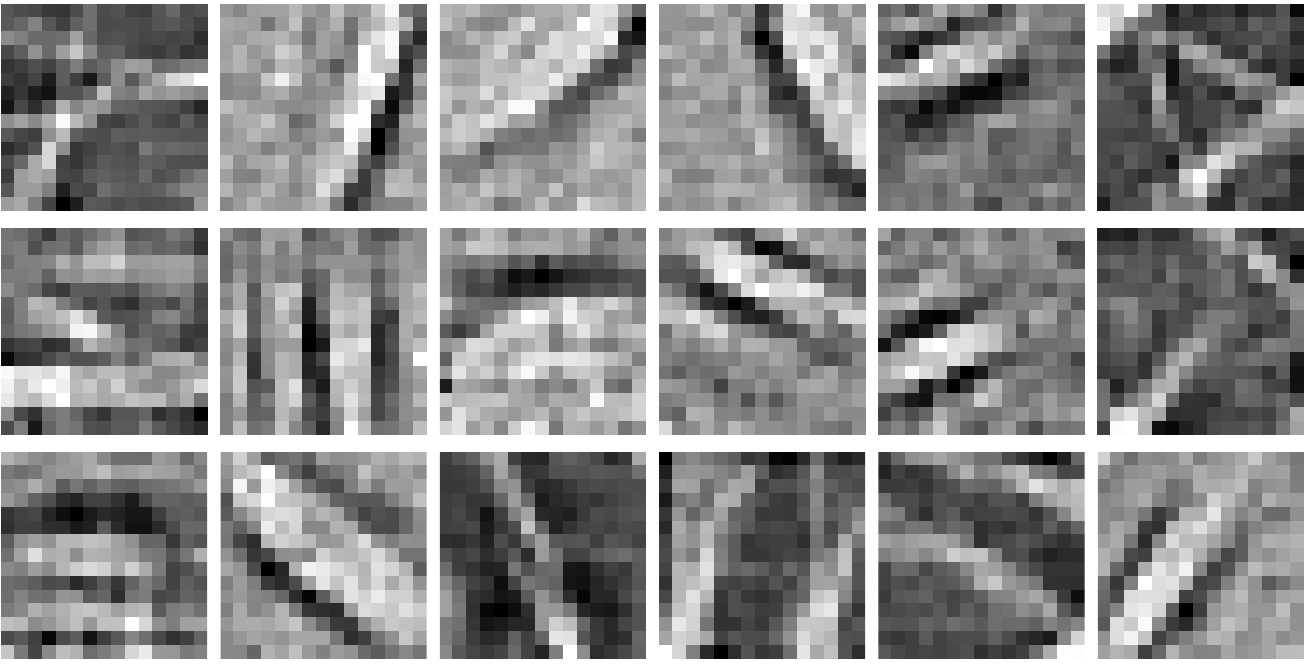}
\includegraphics[width=18em]{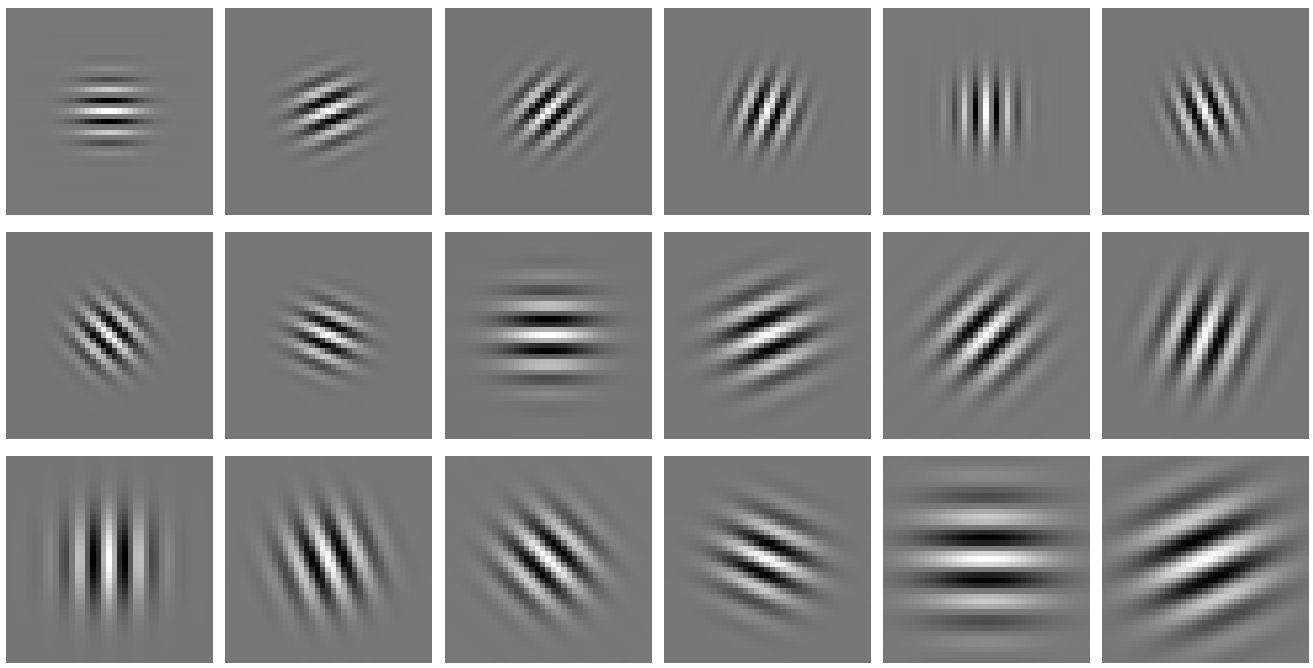}
\caption{Visualisation of the learned filters. Left: randomly selected filters from the first layer in our model; right: the real parts of some Gabor filters.}\label{fig:f}
\end{figure}

\vspace{0.1cm}
\noindent\textbf{Qualitative Results:}\quad Figure~\ref{fig:qualitative} shows some qualitative results. Some examples of surprisingly tough successes are shown in green. Mistakes made by the network (red) (intended category of the sketcher in black) are very reasonable. The clear challenge level of their ambiguity demonstrates why reliable sketch-based communication is hard even for humans.

\vspace{0.1cm}
\noindent\textbf{What Has Been Learned by Sketch-a-Net:}\quad As illustrated Fig.~\ref{fig:f}, the filters in the first layer of Sketch-a-Net (Fig.~\ref{fig:f}(left)) learn something very similar to the biologically plausible Gabor filters (Fig.~\ref{fig:f}(right)) \cite{gabor1946theory}. This is interesting because it is not obvious that learning from sketches should produce such filters, as their emergence is typically attributed to learning from the statistics of natural images \cite{olshausen1996natural_sparse,StollengaMGS14}.

\vspace{0.1cm}
\noindent\textbf{Running cost:}\quad Our Sketch-a-Net model was implemented using Matlab based on the MatConvNet \cite{Chatfield14} toolbox. We trained our 5-network ensemble for 230 epochs each, with each instance undergoing random data augmentation during each iteration.  This took roughly 80 hours in total on a 2.60GHz CPU (without explicit parallelisation), or 10 hours using a NVIDIA K40-GPU. Note that this means Sketch-a-Net was not trained for long enough to use the full pool of available data augmentations. 

\vspace{0.1cm}
\noindent\textbf{Reproducibility:}\quad For reproducibility and to support future research, our training and testing pipeline is made available at \url{http://www.eecs.qmul.ac.uk/\~tmh/}. 

\vspace{-0.1cm}
\section{Conclusion}\label{sec:con}
\vspace{-0.1cm}
We have proposed a deep neural network based sketch recognition model, which we call Sketch-a-Net, that beats human recognition performance by 1.8\% on a large scale sketch benchmark dataset. Key to the superior performance of our method lies with the specifically designed network model that accounts for unique characteristics found in sketches that were otherwise unaddressed in prior art. The learned sketch feature representation could benefit other sketch-related applications such as sketch-based image  retrieval and automatic sketch synthesis, which could be interesting venues for future work.

\vspace{0.1cm}
\noindent\textbf{Acknowledgements:}\quad This project has received funding from the European Union's Horizon 2020 research and innovation programme under grant agreement No 640891. We gratefully acknowledge the support of NVIDIA Corporation for the donation of the GPUs used for this research.

\bibliography{sketchbib,yyang}
\end{document}